\documentstyle[11pt, colacl]{article}

\title{Word Sense Disambiguation using Optimised Combinations of
  Knowledge Sources}
\author{Yorick Wilks \and Mark Stevenson\\Department of Computer
  Science,\\University of Sheffield,\\Regent Court, 211 Portobello
  Street,\\
Sheffield, S1 4DP\\
United Kingdom\\
{\tt \{yorick, marks\}@dcs.shef.ac.uk}}

\begin{document}
\pagestyle{empty}
\maketitle
\bibliographystyle{acl} 

\begin{abstract}
Word sense disambiguation algorithms, with few exceptions,
  have made use of only one lexical knowledge source.  We describe a
  system which performs unrestricted word sense disambiguation (on all
  content words in free text) by combining different knowledge
  sources: semantic preferences, dictionary definitions and
  subject/domain codes along with part-of-speech tags. The usefulness
  of these sources is optimised by means of a learning algorithm. We
  also describe the creation of a new sense tagged corpus by combining
  existing resources. Tested accuracy of our approach on this corpus
  exceeds 92\%, demonstrating the viability of all-word disambiguation
  rather than restricting oneself to a small sample.
\end{abstract}

\section{Introduction}

This paper describes a system that integrates a number of partial 
sources of information to perform word sense disambiguation (WSD) of 
content words in general text at a high level of accuracy.

Word sense disambiguation has become an established, separate, NLP
task, and a module within more general systems to perform useful tasks
like information extraction (IE)\cite{Paz97}.  However, it is still
not a task with an agreed methodology or evaluation criterion, as we
shall discuss below.  Moreover, there is no real evidence as yet that
WSD is useful for practical, general, tasks like IE and machine
translation (MT); the hunch that it is comes largely from the folk
memory that word-sense ambiguity was a major barrier to MT. We are
optimistic and already applying the system described here to IE within
the ECRAN project (see below), but the fact that ECRAN's IE
performance figures improve with WSD will not, of course, be proof
that WSD should get the credit.  The source of scepticism comes from
the old central AI tradition: that all NLP tasks, including WSD, are
knowledge-dependent and therefore cannot be modularised and solved in
isolation by a range of semantic and syntactic considerations of the
sort investigated here.  The problem with this line of argument is
that it tells equally against every NLP task that has been
successfully modularised, and evaluated at a high level of success,
right down to part-of-speech tagging, about which many in the AI
community (one of the present authors included) were sceptical fifteen
years ago.

The methodology and evaluation of WSD are somewhat different from
those of other NLP modules, and one can distinguish three aspects of
this difference, all of which come down to evaluation problems, as
does so much in NLP these days.  First, researchers are divided
between a general method (that attempts to apply WSD to all the
content words of texts, the option taken in this paper) and one that
is applied only to a small trial selection of texts words (for example
\cite{Schutze92} \cite{Yar95short}).  These researchers have obtained
very high levels of success in the mid-to-high ninety percents, close
to the figures for other ``solved'' NLP modules, the issue being whether
these small word sample methods and techniques will transfer to
general WSD over all content words.

Others, \cite{Wilks97short} \cite{Mah97} \cite{Har97} have pursued the
general option on the grounds that it is the real task and should be
tackled directly, but with rather lower success rates.  The division
between the approaches probably comes down to no more than the
availability of gold standard text in sufficient quantities, which is
more costly to obtain for WSD than other tasks.  In this paper we
describe a method we have used for obtaining more test material by
transforming one resource into another, an advance we believe is
unique and helpful in this impasse.

However, there have also been deeper problems about evaluation, which
has led sceptics like \cite{Kil93} to question the whole WSD
enterprise, for example that it is harder for subjects to assign one
and only one sense to a word in context (and hence the produce the
test material itself) than to perform other NLP related tasks. We have
discussed Kilgarriff's figures elsewhere \cite{Wilks97b} and argued that
they are not, in fact, as gloomy as he suggests, and certainly other
researchers have achieved perfectly reasonable levels \cite{Green89}
of inter-subjective agreement to be a basis for this task. Again, this
is probably an area where there is an ``expertise effect'': some
subjects can almost certainly make finer, more inter-subjective, sense
distinctions than others in a reliable way, just as lexicographers do.

But there is another, quite different, source of unease about the
evaluation base: everyone agrees that new senses appear in corpora
that cannot be assigned to any existing dictionary sense, and this is
an issue of novelty, not just one of the difficulty of discrimination.
If that is the case, it tends to undermine the standard
mark-up-model-and-test methodology of most recent NLP, since it will
not then be possible to mark up sense assignment in advance against a
dictionary if new senses are present.  One of us has argued elsewhere
\cite{Wilks97b} that the situation may not be serious, even if such
novelty is frequent, since a subject assigning the ``closest'' available
sense may be sufficient for reliable experiments.  We shall not tackle
this difficult issue further here, but press on towards experiment.

\section{Knowledge Sources and Word Sense Disambiguation}

One further issue must be mentioned, because it is unique to WSD as a task 
and is at the core of our approach.  Unlike other well-known NLP modules, 
WSD seems to be implementable by a number of apparently different 
information sources.  All the following have been implemented as the basis 
of experimental WSD at various times: part-of-speech, semantic preferences, 
collocating items or classes, thesaural or subject areas, dictionary 
definitions, synonym lists, among others (such as bilingual equivalents in 
parallel texts).  These phenomena seem different, so how can they all be, 
separately or in combination, informational clues to a single phenomenon, 
WSD? This is a situation quite unlike syntactic parsing or part-of-speech 
tagging: in the latter case, for example, one can write a Cherry-style rule 
tagger or an HMM learning model, but there is no reason the believe 
these represent different types of information, just different ways of 
conceptualising and coding it.  That seems not to be the case, at first 
sight, with the many forms of information for WSD.  It is odd that this has 
not been much discussed in the field.

In this work, we shall adopt the methodology first explicitly noted in
connection with WSD by \cite{McR92}, and more recently \cite{Ng96short},
namely that of bringing together a number of partial sources of
information about a phenomenon and combining them in a principled
manner.  This is in the AI tradition of combining ``weak'' methods
for strong results (usually ascribed to Newell \cite{New73}) and used
in the CRL-NMSU lexical work on the Eighties \cite{Wilks90}.  We
shall, in this paper, offer a system that combines the three types of
information listed above (plus part-of-speech filtering) and, more
importantly, applies a learning algorithm to determine the optimal
combination of such modules for a given word distribution; it being
obvious, for example, that thesaural methods work for nouns better
than for verbs, and so on.

\section{The Sense Tagger}

We describe a system which is designed to assign sense tags from a
lexicon to general text. The lexicon we chose was the {\it Longman
  Dictionary of Contemporary English} (LODCE)
, which has been used extensively in machine-readable dictionary
research (eg.  \cite{Wilks90}, \cite{Cowie92short},
\cite{Bruce92short}). The senses for each word in LDOCE are grouped
into {\it homographs}, sets of senses with related meanings.

\subsection{Preprocessing}

Before the filters or partial taggers are applied the text is
tokenised, lemmatised, split into sentences and part-of-speech tagged
using the Brill part-of-speech tagger \cite{Brill92}.  A named entity
identifier is then run over the text marking and categorising proper
names.  

Our system disambiguates only the content words in the
text\footnote{We define content words as nouns, verbs, adjectives and
  adverbs, prepositions are not included in this class.} (the
part-of-speech tags assigned by Brill's tagger are used to decide
which are content words) and does not attempt to disambiguate any of
the words which were identified as part of a named entity. For each of
the words being the disambiguated, the system retrieves each of its
possible senses from LDOCE and stores them with the word.

\subsection{Part-of-speech}\label{sec:pos}

Previous work by \cite{Wil98} has shown that part-of-speech tags can
play an important role in the disambiguation of word senses. They
carried out a small experiment on a 1700 word corpus taken from the
{\it Wall Street Journal} and, using only part-of-speech tags,
attempted to find the correct LDOCE homograph for each of the content
words in the corpus. They part-of-speech tagged the text using Brill's
tagger and removed from consideration any homograph whose
part-of-speech category did not agree with the tags assigned by
Brill's system.  They then chose the most frequently occurring
homograph of the remaining homographs as the tag for that word. They
found that 92\% of content words were assigned the correct homograph
compared with manual disambiguation of the same texts.

While this method will not help us disambiguate within the homograph, since 
all senses which combine to form an LDOCE homograph have the same 
part-of-speech, it will help us to identify the senses completely 
innapropriate for a given context (when the homograph's part-of-speech 
disagrees with that assigned by a tagger).  

It could be reasonably argued that this is a dangerous strategy since,
if the part-of-speech tagger made an error, the correct sense could be
removed from consideration.  As a precaution against this we have
designed our system so that if none of the dictionary senses for a
given word agree with the part-of-speech tag then they are all kept
(none removed from consideration).  

\subsection{Dictionary Definitions}\label{sec:sim_anneal}
 
\cite{Lesk86} proposed a method for sense disambiguation using overlap
of the dictionary definitions of words as a measure of their semantic
closeness. In this way it is possible,
at least in theory, to tag each word in a sentence with its sense from
any dictionary which contains textual definitions for its senses.
However, it was found that the computations which would be necessary
to test every combination of senses, even for a sentence of modest
length, was prohibitive.
 
\cite{Cowie92short} used simulated annealing to optimise Lesk's
algorithm. By applying this method to Lesk's heuristic Cowie et.  al.
found that they made the process of optimising the sense choice
tractable, often choosing an assignment of senses from as many as
$10^{10}$ choices.  The optimisation was over a simple count of words
in common in definitions, however, this meant that longer definitions
were preferred over short ones, since they have more words which can
contribute to the overlap, and short definitions or definitions by
synonym were correspondingly penalised.  We attempted to solve this
problem as follows.  Instead of each word contributing one we
normalise its contribution by the number of words in the definition it
came from.  The Cowie et. al. implementation returned one sense
for each ambiguous word in the sentence, without any indication of the
system's confidence in its choice, but, we have adapted the
system to return a set of suggested senses for each ambiguous word in
the sentence.  We found that the improved
evaluation function led to an improvement in the algorithm's
effectiveness.

\subsection{Pragmatic Codes}

Our next partial tagger is based on the technique of disambiguation by
examining thesaural hierarchies (such as Roget 
and WordNet
). LDOCE contains a hierarchy of subject codes which indicate the
subject of a text: these are divided into primary, of which there are
around 300, and secondary codes (around 2,500). The pragmatic code
associated with a sense consists of four letters, for example {\tt
  ECZA}, the first two letters of which indicate the primary code
(economics in this case) and the final two the sub-class of the
primary code (accounting in this case). The hierarchy is therefore
shallow (since there are only two levels) but wide (since there are
many subject codes at each level).
 
We carried out disambiguation using a modified version of the
simulated annealing algorithm, which attempts to optimise the number
of pragmatic codes of the same type in the sentence.  However, the
method has been modified from the application to word overlap in three
ways: first, we maximise the overlap of the pragmatic codes associated
with the word senses rather than the content words in their
definitions.  Secondly, we optimise over entire paragraphs rather than
just sentences, because there is good evidence \cite{Gale92} that a
wide context, of around 100 words, is optimal when disambiguating
using domain codes.  Finally, we only optimise the pragmatic codes for
nouns, since \cite{Yar93} has shown that, in general, nouns are
disambiguated by ``broad context'' considerations, such as the general
subject of the text they are in, while other parts of speech are
disambiguated by ``local context'', such as the semantic types of the
words they modify.

\subsection{Selectional Restrictions}\label{sect:selectional}
 
LDOCE senses contain simple selectional restrictions for each content
word in the dictionary. A set of 35 semantic classes are used,
such as {\tt H} = Human, {\tt M} = Human male, {\tt P} =
Plant, {\tt S} = Solid and so on. Each word sense for a noun is given
one of these semantic types, senses for adjectives list the type which
they expect for the noun they modify, senses for adverbs the type they
expect of their modifier and verbs list between one and three types
(depending on their transitivity) which are the expected semantic
types of the verb's subject, direct object and indirect object.
 
We identify grammatical links between verbs, adjectives and adverbs 
and the head noun of their arguments using a specially constructed shallow 
syntactic analyser.
 
The semantic classes in LDOCE are not provided with a hierarchy,
but, Bruce and Guthrie \cite{Bruce92short} manually identified
hierarchical relations between the semantic classes, constructing them
into a hierarchy which we use to resolve the restrictions.
 
The selectional restriction resolution algorithm makes use of the
information provided by the shallow syntactic analyser and the named
entity identifier.  Although we are not disambiguating named entities
they are still useful to help disambiguate other words: for example,
if a verb has two senses one of which places the restriction {\tt H}
(=Human) on its object, the other {\tt I} (=Inanimate) and the object
was a named entity marked {\tt PERSON} then we would prefer the first
sense.  Restrictions are resolved by returning all the senses which
agree with their restrictions (that is, those whose semantic category
is at the same, or a lower, level in the hierarchy).

\section{Combining Knowledge Sources}

Since each of our partial taggers suggests only possible senses for
each word it is necessary to have some method to combine their
results.  We trained decision lists \cite{Riv87} using a supervised
learning approach.  Decision lists have already been successfully
applied to lexical ambiguity resolution in \cite{Yar95short} where
they perfromed well.

We present the decision list system with a number of training words for 
which the correct sense is known.  For each of the words we supply each of 
its possible senses (apart from those removed from consideration by the 
part-of-speech filter (Section \ref{sec:pos})) within a context consisting 
of the results from each of the partial taggers, frequency information and 
10 simple collocations (first noun/verb/preposition to the left/right and 
first/second word to the left/right).  Each sense is marked as either {\tt 
appropriate} (if it is the correct sense given the context) or {\tt 
inappropriate}.  A learning algorithm infers a decision list which 
classifies senses as {\tt appropriate} or {\tt inappropriate} in context.  
The partial taggers and filters can then be run over new text and the 
decision list applied to the results, so as to identify the appropriate 
senses for words in novel contexts.

Although the decision lists are trained on a fixed vocabulary of words
this does not limit the decision lists produced to those words, and
our system can assign a sense to any word, provided it has a
definition in LDOCE.  The decision list produced consists of rules
such as ``if the part-of-speech is a noun and the pragmatic codes
partial tagger returned a confident value for that word then that
sense is appropriate for the context''.

\section{Producing an Evaluation Corpus}

Rather than expend a vast amount of effort on manual tagging we
decided to adapt two existing resources to our purposes.  We took
SEMCOR
, a 200,000 word corpus with the content words manually tagged as part of
the WordNet project. The semantic tagging was carried out under
disciplined conditions using trained lexicographers with tagging
inconsistencies between manual annotators controlled.  SENSUS
\cite{Kni94short} is a large-scale ontology designed for
machine-translation and was produced by merging the ontological
hierarchies in WordNet and LDOCE \cite{Bruce92short}.  To facilitate
this merging it was necessary to derive a mapping between the senses
in the two lexical resources.  We used this mapping to translate the
WordNet-tagged content words in SEMCOR to LDOCE tags.
 
The mapping is not one-to-one, and some WordNet senses are mapped onto
two or three LDOCE senses when the WordNet sense does not distinguish
between them. The mapping also contained significant gaps (words and senses
not in the translation). SEMCOR contains 91,808 words tagged with
WordNet synsets, 6,071 of which are proper names which we ignore,
leaving 85,737 words which could potentially be translated.  The
translation contains only 36,869 words tagged with LDOCE senses,
although this is a reasonable size for an evaluation corpus given this
type of task (it is several orders of magnitude larger than those used
by \cite{Cowie92short} \cite{Har97} \cite{Mah97}).  This corpus was
also constructed without the excessive cost of additional hand-tagging
and does not introduce any inconsistencies which may occur with a
poorly controlled tagging strategy.

\section{Results}

To date we have tested our system on only a portion of the text we
derived from SEMCOR, which consisted of 2021 words tagged with LDOCE
senses (and 12,208 words in total). The 2021 word occurances are made
up from 1068 different types, with an average polysemy of 7.65.  As a
baseline against which to compare results we computed the percentage
of words which are correctly tagged if we chose the first sense for
each, which resulted in 49.8\% correct disambiguation.

We trained a decision list using 1821 of the occurances (containing
1000 different types) and kept 200 (129 types) as held-back training
data. When the decision list was applied to the held-back data we
found 70\% of the first senses correctly tagged. We also found that
the system correctly identified one of the correct senses 83.4\% of
the time. Assuming that our tagger will perform to a similar level
over all content words in our corpus if test data was avilable, and we
have no evidence to the contrary, this figure equates to 92.8\%
correct tagging over all words in text (since, in our corpus, 42\% of
words tokens are ambiguous in LDOCE).

Comparative evaluation is generally difficult in word sense
disambiguation due to the variation in approach and the evaluation
corpora. As \cite{Res97} said ``there are nearly as many test suites
as there are researchers in the field''.  However, it is fair to
compare our work against other approaches which have attempted to
disambiguate all content words in a text against some standard lexical
resource.  Examples of this are \cite{Cowie92short}, \cite{Har97},
\cite{McR92}, \cite{Ver90short} and \cite{Mah97}.  Neither McRoy nor
Veronis \& Ide provide a quantative evaluation of their system and so
our performance cannot be easily compared with theirs. Mahesh et. al.
claim high levels of sense tagging accuracy (about 89\%) from a single
knowledge source, the Mikrokosmos knowledge representation, but our
results are not directly comparable since its authors explicitly
reject the conventional markup-training-test method used here. Cowie
et.  al.  used LDOCE and so we can compare results using the same set
of senses.  Harley and Glennon used the {\it Cambridge International
  Dictionary of English}
which is a comparable resource containing similar lexical information
and levels of semantic distinction to LDOCE. Our result of 83\%
compares well with the two systems above who report 47\% and 73\%
correct disambiguation for their most detailed level of semantic
distinction.  Our result is also higher than both systems at their
most rough grained level of distinction (72\% and 78\%).  These
results are summarised in Table \ref{tab:compare}.

\begin{table*}[!htbp]
\begin{center}
\begin{tabular}{|c|c|c|c|}
\hline
System & Resource & Ambiguity level & Result\\
\hline
\hline
& & homograph & 72\% \\ \cline{3-4}
\raisebox{1.5ex}[0pt]{\cite{Cowie92short}} & \raisebox{1.5ex}[0pt]{LDOCE} & sense & 47\%\\
\hline
& & `coarse' level & 78\% \\ \cline{3-4}
\raisebox{1.5ex}[0pt]{\cite{Har97}} & \raisebox{1.5ex}[0pt]{CIDE} & `fine' level & 73\%\\
\hline
Reported system & LDOCE & sense & 83\%\\
\hline
\end{tabular}
\caption{Comparison of tagger with similar systems}\label{tab:compare}
\end{center}
\end{table*}

In order to compare the contribution of the separate taggers we
implemented a simple voting system.  By comparing the results obtained
from the voting system with those from the decision list we get some
idea of the advantage gained by optimising the combination of
knowledge sources.  The voting system provided 59\% correct
disambiguation, at identifying the first of the possible senses, which
is little more than each knowledge source used separately (see Table
\ref{tab:res}).  This provides a clear indication that there is a
considerable benefit to be gained from combining disambiguation
evidence in an optimal way. In future work we plan to investigate
whether the apparently orthogonal, independent, sources of information
are in fact so.

\begin{table*}[!htb]
\begin{center}
\begin{tabular}{|c|c|c|}
\hline
Knowledge Sources & Result\\
\hline
\hline
Dictionary definitions & 58.1\% \\
\hline
Pragmatic codes & 55.1\% \\
\hline
Selectional Restrictions & 57\% \\
\hline
All & 59\% \\
\hline
\end{tabular}
\caption{Results from different knowledge sources}\label{tab:res}
\end{center}
\end{table*}

\section{Conclusion}

Our conclusion is that all content words can be disambiguated as a
close to acceptable level by an optimised combination of lexical
knowledge sources and a part-of-speech filter: 92\% of all words in
text can be disambiguated in this way.

\vspace{-2pt}

\section*{Acknowledgments}
\vspace{-2pt}
{\small The work described in this paper has been supported by the European
Union Language Engineering project ``ECRAN -- Extraction of Content:
Research at Near-market'' (LE-2110).}

\vspace{-10pt}


\end{document}